# Explaining AI as an Exploratory Process: The Peircean Abduction Model

September 2020

R.R. Hoffman
William J. Clancey
Institute for Human and Machine Cognition

Shane T. Mueller
Michigan Technological University

[rhoffman@ihmc.us]

## Abstract

Current discussions of "Explainable AI" (XAI) do not much consider the role of abduction in explanatory reasoning (see Mueller, et al., 2018). It might be worthwhile to pursue this, to develop intelligent systems that allow for the observation and analysis of abductive reasoning and the assessment of abductive reasoning as a learnable skill. Abductive inference has been defined in many ways. For example, it has been defined as the achievement of insight. Most often abduction is taken as a single, punctuated act of syllogistic reasoning, like making a deductive or inductive inference from given premises. In contrast, the originator of the concept of abduction—the American scientist/philosopher Charles Sanders Peirce—regarded abduction as an exploratory activity. In this regard, Peirce's insights about reasoning align with conclusions from modern psychological research. Since abduction is often defined as "inferring the best explanation," the challenge of implementing abductive reasoning and the challenge of automating the explanation process are closely linked. We explore these linkages in this report. This analysis provides a theoretical framework for understanding what the XAI researchers are already doing, it explains why some XAI projects are succeeding (or might succeed), and it leads to design advice.

## Explanation as an Exploratory Process: The Peircean Abduction Model

> *"Precisely how much of the business of thinking a machine could possibly be made to perform, and what part of it must be left for the living mind, is a question not without considerable practical importance."* C.S. Peirce, 1887, p. 165

## Introduction

Abductive inference remains a significant challenge for the field of artificial intelligence. Indeed, it is possible to formalize a number of classes of abductive problems (e.g., abductions involving premises that have no incompatibilities). (For a review of formal models of abduction see Bylander, et al., 1991). Many philosophers and psychologists have discussed abduction (hypothesis formation and testing) as basic to scientific reasoning (e.g., Bruner, 1985; Collins, 1985; Glaser, 1984; Haig, 2005; Pfeiffer, et al., 1987; Nummedal, 1987; Schurz, 2008; Selz, 1935). Abduction has been cited as an aspect of, or even a central aspect of critical thinking, cognitive flexibility, fluid intelligence, and creativity (e.g., Douven, 2011; Moore and Malinowski, 2009).

Diverse definitions of abduction can be found  in the pertinent literatures. Some reports consider abduction as "not a logically sound deduction or proof." In some reports, what is referred to as abduction is actually a form of induction (see Lenat, 2013, p. 54). This includes inductive generalizations from sets of cases in case-based reasoning. Abduction has been defined as the achievement of "insight" (e.g., Hoffmann, 2010, 2011). Abduction has been defined variously as an accomplishment, an ability, a skill, and as an amalgam of component skills such as evaluating evidence, forming mental models, recognizing cues, or other hypothetical cognitive processes. Scholars have proposed various patterns of abductive explanation (as many as 15 of them), such as "existential abduction, "factual abduction," "analogic abduction," "diagrammatic abduction," and "theoretic abduction" (Schurz, 2008). It makes sense that there are indeed distinguishable patterns of explanatory arguments. Different things can be inferred (e.g., concepts, laws, models, belief systems) and things can be inferred in different ways (e.g., by selecting hypotheses from memory or by creating new hypotheses). But such elaborations and taxonomies have triggered considerable disagreement about what does and does not count as abduction.

In situations such as this we like to start by looking at the seminal definition and anchor the discussion there. Our goal in this essay is to attempt some clarification by revisiting Charles Sanders Peirce's original description, and then trying to see how far modern scientists have gone in implementing it. Is abduction a single, punctuated act of reasoning, like making a logical inference from given premises? Is the property of "being an explanation" a property of propositions? Since abduction is often defined as "inferring the best explanation," these two questions are closely linked. We explore them in this essay.

We begin by talking a theoretical/scientific perspective, centered (unabashedly) on the community of philosophers, psychologists, and computer scientists who write and think about abduction. We then attempt to make contact with the world of Explainable AI (XAI) and Intelligent Tutoring Systems (ITS), culminating in some extrapolations in the concluding section.

**Peirce the Logician**

The concept we call abduction might actually be traced to Aristotle. In *Prior Analytics* (69a 20ff; Smith, 1989), Aristotle discussed what he called "reduction"—the transformation of the subject-predicate relations in the major and minor premises of a syllogism (e.g., Socrates is mortal, Socrates is a man), so that each of the four basic types (or "figures") of syllogisms could be expressed as a type in which the conclusion is an explanatory rule, or we might say, an assertion about a class (e.g., All men are mortal). Despite this historical precedent, most discussions attribute the concept of abduction to the early American philosopher/psychologist/logician Charles Sanders Peirce.

In classical formal logics, deduction and induction are completely defined as sets of a few assertions, one of which must be a generalization (that is, an assertion about a class). In contrast, abduction depends on the reasoner's knowledge—propositions that come from beyond the given the observation(s) and premises. Hence, abduction is not the same as inductive enumeration. Nonetheless, to philosophers abduction is a form of inference (Douven, 2011; Harman, 1956; Lipton, 2004).

In writings spanning 1867-1902, Peirce referred to abduction as "hypothetic inference"—the inferring of a hypothetical explanation from an observed circumstance. More than this, abduction involves justification. A proposed explanation is "best" because if it were correct the observed circumstance would necessarily occur. In this view, abduction would be distinct from both deduction and induction. Indeed, it would be a third basic or fundamental type of inference. This would seem to retain abduction in the mold of classical logic.

In his discussion of the concept of logical machines, Peirce (1887) restricted himself primarily to the conduct of syllogistic inferences in machines designed to go from selected sets of affirmative and negative premises. But Peirce thought of inference in a way that differed from the classical logics of causation of scholars such as Aristotle and David Hume (1740, Part III). Peirce did not seem to mean that abduction was a third classical type (in addition to deduction and induction), rather, to Peirce one single type covers the classical forms.

> RULE (All men are mortal)
> Case 1 (Socrates is a man)
> _______________________________
> Case 1 falls under the rule (Socrates is mortal).

Induction is looking at a number of cases and having an expectation of what you would find for future cases. Thus, it can be said that induction is the same thing as generalization, and it is always relative to some bounded set or limiting parameters. Of course the difficulty here is that since any given case will have an unbounded number of hypothetical features (potential premises) about which one might form rules, one can have an unbounded number of rules (see Douven, 2011). While this is a distinct problem for logic, it does not seem to have been a concern for the Peirce the pragmaticist.

Recent philosophy has maintained this linkage to classical logic. To Gary Shank (1998, p.847-848), abduction is "rendering what might be thought of as a unique experience into an instance of a more general phenomenon." John and Susan Josephson and Josephson (1995) also described abduction in a formal way (see also Aliseda, 2007; Haig, 2005; Minnameier, 2010):

> D is a collection of facts, observations, or "givens,"
> H explains D, that is, H would, if true, explain D,
> No other hypothesis can explain D as well as H does,
> Therefore, H is probably true.

In this definition we find an explicit linkage of the fuzzy concept of abduction to the fuzzy concept of explanation.

Peirce the logician presented a stronger claim, asking whether abduction is really a third fundamental type of inference. In some of his discussions of abduction (e.g., 1903) he seems to have considered it as a hybrid, that is, abductive reasoning episodes also involve induction and deduction. As Peirce put it, "abduction can partake of the nature of induction" (CP 6.522-8).

But Peirce's definition was psychological (i.e., reliant on the notion of explanation) as well as logical (representing abductive inferences formally). To Peirce, the match between a set of data and a preferred explanatory frame is more plausible than the match to some other explanatory frame or frames, and so we accept the preferred frame as the likely explanation. An abductive inference (an hypothesis) is maintained until contradicted by experience, or until experience suggests a better (simpler, more general, more plausible, etc.) hypothesis.

The following Table integrates the things Peirce said about abduction in his various writings (see also Fann, 1970).

Table 1.  The full Peircean concept of abduction.

| PROCESS | REQUIREMENTS |
|---|---|
| 1. Observation of an event or phenomenon. | The observed event or phenomenon is interesting or surprising. <br><br> The perception of the event or phenomenon (i.e., categorization) hinges on the reasoner's knowledge and concepts. <br><br> Discussions in the literature often refer to relatively simple examples of abduction, but it is clear especially in Peirce's writing that there is an assumption that the observed event or phenomenon is at least non-trivial, that is, it is a complex event or phenomenon. i.e., it is system of interactions whose properties (structure/behaviors) are not understood. |
| 2. Generation of one or more possible explanations for some observed event or phenomenon. | The understanding of the event or phenomenon hinges on the reasoner's knowledge and concepts (a sensemaking process). <br> The search for a preferred explanation presupposes that the reasoner has more than one candidate explanation in mind. |
| 3. Judging the plausibility of the candidate explanation(s). | The judgment can be but is not necessarily based on rationalist considerations of necessity and sufficiency. <br><br> The judgment can be but is not necessarily based on the estimation of probabilities or likelihoods. |
| 4. Resolving the explanation. | The plausibility judgment typically (though not necessarily) results in a determination that a particular explanation is preferred. <br><br> "Inference to the best explanation" (Harman, 1965) implies that one has in mind more than one candidate explanation or "competing |
| | theory" (see row 2, above) and all but one of them is rejected Proctor (2008). |
| 5. Extending the explanation. | The determination of a preferred explanation is always tentative, that is, subject to disconfirmation by further inquiry even though there is an assumption that further instances will conform to the preferred explanation. |

It is noteworthy that these requirements are consistent with the logic-inspired views of Harman (1965), Josephson and Josephson (1995), and Lipton (2004), and the psychology-inspired views of Proctor (2008) and Lombrozo (2012). "The process of explaining recruits prior beliefs and a host of explanatory preferences, such as unification and simplicity, that jointly constrain subsequent processing" (Lombrozo, 2012, p. 260). This suggests a reconciliation of the logical view and the psychological view.

**Reconciling the Logical and the Psychological**

There is a tendency for discourses on abduction to either define it  philosophically in terms of other fuzzy concepts (e.g., critical thinking, explanation, etc.) or to define it logically as a form of inference based on a set of premises. But there is an alternative, in which abduction is understood as exploratory modelling; a constructive reasoning activity that is highly dependent on knowledge. Tania Lombrozo, a leading psychologist in the area of explanatory reasoning, regards explanation and abduction as separate phenomena: And as we pointed out above, the reliance on propositions that are external to the abductive syllogism (no matter how many premises it begins with) is central in the Peircean definition of abduction.

Referring to the Table 1, the reasoner's motivation for creating a model is that the observed event or phenomenon is interesting or surprising. And the observed event or phenomenon is at least non-trivial, that is, it is complex and involves interactions whose properties (structure/behaviors) are not understood. To Peirce, abduction is an accomplishment, a name for an exploratory activity that begins with some information (data, assumptions, hypotheses, partial models, etc.) about some system of interest and produces a more complete model of that system, one that is consistent with the given information.

In this exploratory modelling view, one might avoid invoking the concept of explanation entirely. This is because an explanation is, after all, itself a modelling activity. One might say something like "If H is a correct characterization (model) of the system of interest, then the collection or assertions D is logically consistent with H's structure/processes, i.e., D is modeled by H; H models D."

More important for present purposes is Peirce's notion that abduction as an activity involves going beyond model formation to active exploration. This allows for the discovery and description of new or novel experiences, and the rule is tested for plausibility by characterizing the new experiences. As a search for a satisfying explanation, abduction has an aesthetic component or aspect. As a search for a satisfying explanation, abduction has the aspect of a plausibility judgment. Abduction is an activity that is extended in time, having its own structure and processes, often requiring representational tools. It is not a single act of "reasoning" like making a logical inference.

To Peirce the abductive derivation of a rule is a creative act. Hence, we come full circle to consider the relation of abduction to other psychological processes, such as sensemaking, creativity, and critical thinking.

## Abduction and Sensemaking

As is indicated in row 2 in Table 1, the notion that abduction is inference to the "best" hypothesis presumes that the reasoner has more than one hypothesis in mind. But on this matter, Peirce himself muddied the water. Peirce referred to the phenomenon in which an explanation pops out in the reasoner's awareness. "...the suggestion comes to us like a flash" (Peirce, 1891; (CP 5.542, 5.544-5, 5-157). In other words, the very act of perceiving something—the categorizing of it—is a type of hypothesis formation, albeit an extreme case, as Peirce says. Is that moment to be regarded as an act of inference? If so, it is not a classical form. In contrast, it hearkens to insight, or the "Eureka" phenomenon. In psychological theory, this is "recognition-primed decision making" (Klein, 1993), in which the expert rapidly apprehends a situation and immediately conceives the best possible course of immediate action without any deliberation over alternative courses of action. There is no form of utility analysis.

So, while abduction to a "best" explanation" forces a consideration of the evaluative comparison of explanations—in the logical approach—it does not force it in the psychological approach. Neither recognition of an anomaly nor the evaluation of plausibility requires that abduction is solely an act of recognition or the realization of a *single* explanation. It is the use of the word "best" that implies that more than one explanation is in mind. Peirce did not use that word. He referred to observations that are "interesting" or surprising."

That said, in the full Peircean model there is a consideration of more than one possible causal model, and an evaluation or comparison of those alternative models leads to a plausibility judgment, (i.e., one of them is "best")  In other words, abduction must be more often like sensemaking than recognition-primed decision making, even though the formation of an initial mental model might be very rapid.

To pursue this more specifically, the Peircean model of abduction is consistent with the Data/Frame model of sensemaking  (Klein, et al., 2006), which derived from studies of domain experts. The D/F model asserts that sensemaking involves the formation of an initial mental model, or frame. The frame raises questions and allows the reasoner to know when to recognize anomalies. The frame is evaluated for its plausibility and is explored in a search for confirming and disconfirming evidence; it might be adjusted, or discarded altogether. Referring to Table 1:
- Abduction involves the observation of something that is surprising (row 1).
- Abduction depends on the reasoner's knowledge, not just on a set of given premises (row 2).
- Abduction involves a plausibility judgment (rows 3 and 4).
- The abduction is extended by a search for additional information (row 5).

The key difference between the Peircean model and the Data/Frame model is that the Peircean model seems more like a linear narrative, whereas the Data/Frame model is entirely closed loop (the data suggest a frame, but at the same time, the frame determines what counts as additional data). This difference aside, it is noteworthy that modern research on "naturalistic decision making" has led to models that are consistent with Peirce's model of abduction.

This includes consistency with the "Plausibility Cycle" model of self-explanation (Klein, et al., 2019). The premise of XAI is that the XAI system would generate explanations and present them to the user. Psychological research on causal reasoning about complex "real world" events and complex systems contexts has shed light on how that people actively engage in self-explanation. Some sort of surprising event (or "trigger") leads to a process of story building. But then plausibility gaps are revealed and an attempt is made to fill those gaps. Finally, the explanatory narrative is tested, sometimes leading to insights that resolve the initial surprise.

In other words, from the psychological perspective the full Peircean model of abduction is consistent with models of explanatory reasoning that have emerged in recent years in the field of applied cognitive psychology and the study of naturalistic decision making. These conceptual models are all getting at (more or less) the same things.

Another psychological coherence involves the relation of abduction and critical thinking.

**Abduction and Critical Thinking**

Apparently little research has been done on the skill or capacity for generating plausible best-guess hypotheses (Lombrozo, 2012). That said, some researchers have demonstrated success at teaching critical thinking skills defined in such as way at to embrace the concept of abduction (e.g., Schank, 2011).

Kees van Dongen and colleagues (2003) conducted critical thinking training using an intelligent tutoring system that encouraged trainees to list multiple alternative hypotheses and then list both the confirming and disconfirming evidence for each hypothesis. The manifest purpose of the training tool was to mitigate confirmation bias, but the tutoring involved practice on deciding which of a set of alternative hypotheses was the best hypothesis.

Another study that relied on a conceptual model that approximates the Peircean concept of abduction is one by Karel van den Bosch and Marlous de Beer (2007) on training for decision making. The researchers described two elements of critical thinking:
- *Building a story.* Explaining a situation, integrating assumptions and uncertainties. This corresponds to rows 1 and 2 in Table 1, above.
- *Testing and Evaluating a Story.* Identifying incomplete and contradictory information; evaluating the plausibility of the story. This corresponds to rows 3 and 4 in the Table above.

To Peirce the abductive derivation of a rule is a motivated attempt to understand. Hence, we come full circle to consider the relation of abduction to explanation.

**Abduction and Explanation**

In the field of AI, broadly, it has been recognized for quite some time that explanation is an exploratory process. Hence, it comes as no sur[rise that schemes for explanatory systems hearken to elements of the Peircean model.  For instance, Roger Schank (1986) said:

> *... understanding means finding an extant knowledge structure in memory and placing the information being processed within that structure... Explanations that are critical to learning are those that are intended to add knowledge by the process of explanation... (p. 27). The interesting case is one in which we find ourselves initially failing to comprehend something and then being able to construct what we consider to be a fairly plausible explanation that at least momentarily  leaves us believing that we have now understood what was  at first incomprehensible. In a sense being told an answer can often be a waste of time. the process of constructing an explanation forces learning to occur by  causing certain questions to be generated that may continue to be asked for a long time...* (p. 228).

Schank's analysis includes these Peircean elements: (1) Some sort of surprise or anomaly is observed, (2)  the learner is reminded of some existing knowledge structure that seems relevant, then (3) questions are asked in a deliberative exploratory process.

In the literatures on medical diagnostic systems, abduction has been referenced as the process of inferring causal processes based on observations (e.g., symptoms) (see Finin and Morris, 1998). Similar ideas have been applied in Intelligent Tutoring Systems (ITS) research, in which observed student behaviors are attributed to the student's reasoning and beliefs, often called the program's "student model." Indeed, ITS would seem to be an ideal venue for attempting to develop computational models of abductive reasoning, in the full Peircean sense. However, the fuzziness of the concept of abduction manifests itself in the ITS literature. Indeed, computer scientists have proposed "new forms of abduction" (e.g., Stickel, 1988) that are not obviously tied to a Peircean notion at all.

Although the term abduction is sometimes invoked in research on intelligent systems, it actually does little heavy lifting, either in cognitive modeling or in implementations. For example, in the field of machine learning, Mooney (1998) defined abduction as the process of revising a knowledge base in order to fit the data. He defined abduction formally as the inductive inference across multiple cases of an explanatory hypothesis (cause-effect relation), such that its conjunction with observations of a particular case is true, and its conjunction with background knowledge is also true. Mooney's linkage of abduction to induction accords with Peirce's assertion that "abduction partakes of the nature of induction," but that's as far as it goes.

Another partial alignment with Peircean abduction is the work by Fraser, et al. (1989), who sought to develop an ITS for antibody identification in immunohematology. To do that, they conducted knowledge elicitation interviews with a domain expert. The interviews focused on the recall of critical incidents, in which an initial diagnosis proved wrong.  The researchers' primary intent was to pin diagnostic errors on one or another cognitive bias (e.g., overestimating the likelihood of a given hypothesis).  (While the researchers were able to do this, they also learned that the expert was aware of possible biases, and performed checks on her probabilistic reasoning, by such means as examining base rates.) The expert hematologist's process of inferring a diagnosis based on the evidence available was referred to as abduction, but otherwise the concept of abduction (as an act of inference) played no role in the research. That said, the description of the hematologist's process for testing hypotheses fits with the Peircean model, that is, there was an attempt to find converging or disconfirming evidence: an explicit check on the lab technician's process for performing the test.

The physics problem solving ITS developed by Maxim Makatchev and his colleagues (2004a,b) comes closer to the Peircean model. Given that a learner's mental model of complex systems is likely to be incomplete and inconsistent (DiSessa, 1993, 2018), the ITS of Makatchev, et al. was intended to help students recognize and correct gaps, illogical relations, and misconceptions in their reasoning. In the implementation, student essays about physics problems were rendered as sets of propositions. "[T]he student essay is viewed as a fragmentary, incomplete, and possibly incorrect proof. Our task is to complete that proof insofar as possible." (p.193). The "proofs" were demonstrably coherent arguments that represent the student's mental model. In the implementation, it was the ITS (not the learner) who engaged in abductive reasoning, inferring the best proof in order to guide the tutoring process. That said, the implementation approximates the Peircean scheme in certain respects. First, the implementation included plausibility analysis (by assigning weights for alternative explanations) (see row 3 in Table 1). Second, while it might be asserted that abduction involves inferring explanations from observations, it might also be said that observations are made as a consequence of the explanations (see row 5 in Table 1).

The work by Makatchev et al. illustrates the importance of drawing a distinction between abduction (of a sort) that is the program's process of constructing model of student's reasoning, and abduction by the student in the sense of producing the observed behavior that is being modeled. In 1970s and most of 1980s, the ITS community equated the program's model of the student's knowledge with the student's knowledge (Sleeman and Brown, 1982). While equating a model of the student's knowledge with  the student's mental process may be functionally useful for implementation and testing, with respect to being a psychological explanation of student behavior, it is on shaky grounds. In other words, while we see elements of the concept of abduction in the ITS work—abduction as Peirce described it—there is still a ways to go.

Indeed, according to Bylander, et al. (1991) discovery of a best explanation that is plausible, parsimonious (no subset of the premises is sufficient), and complete (i.e., it explains all the data) is an intractable (NP-hard) problem. *"... choosing between incompatible hypotheses, reasoning about cancellation effects among hypotheses, and satisfying the minimal plausibility requirement are major factors leading to intractability"* (p. 25) (see also Hoffmann, 2011).

## So What?

<u>What are the implications for XAI system design?</u>

The concepts of abduction—ranging from Peirce's philosophical psychology to modern intelligent systems implementations—all speak to the necessity of regarding explanation as an exploratory process. XAI systems that automatically generate "explanations" and then spoon feed them to users might have some value, but it will necessarily be limited. The property of "being an explanation" is not a property of statements, images, displays, or any of that. Rather, it characterizes an activity, usually interactive and sometimes mixed-initiative, with the source information (i.e., the explainer/instructor) asking questions and seeking explanations from the learner as well.

<u>Does the present analysis explain why some XAI projects are succeeding (or might succeed) more than others?</u>

Yes. Most explanations offered by XAI systems are entirely local. They do not explain "how it works"—the methods used by the AI program. They do not allow (or help) the learner to actively explore the boundaries of when and why the AI program works well and when and why it works poorly.

<u>Is this a theoretical framework for understanding what the XAI researchers are already doing?</u>

While some researchers in the area of Explainable AI have understood that the XAI system must possess (or create) a model of the learner's mental model, most XAI research sidesteps this challenge, just as it seems to have sidestepped the decades-old work on intelligent tutoring systems. In practice, it seems a bridge too far to load onto the core challenge of XAI such additional challenges as robust inference over world knowledge, natural language understanding, and the creation of meaningful dialog systems. As a long-term area of research, XAI is a nexus for many of the fundamental problems in the field of AI.

<u>Does the present analysis lead to design advice?</u>

Yes: An explanation capability should include tools to that enable users/learners to discover heuristics for when the AI system works well (and why) and when it doesn't work well (and why). The XAI system should allow users to share their discoveries and form a community in which the self-explaining of individuals can result in sharable knowledge.

### Acknowledgement and Disclaimer

This material is approved for public release. Distribution is unlimited. This research was developed with funding from the Defense Advanced Research Projects Agency (DARPA) under agreement number FA8650-17-2-7711. The U.S. Government is authorized to reproduce and distribute reprints for Governmental purposes notwithstanding any copyright notation thereon.

The views and conclusions contained herein are those of the authors and should not be interpreted as necessarily representing the official policies or endorsements, either expressed or implied, of AFRL or the U.S. Government.